\documentclass[conference]{IEEEtran}

\usepackage{cite}
\usepackage{amsmath,amssymb,amsfonts}
\usepackage{algorithmic}
\usepackage{graphicx}
\usepackage{textcomp}
\usepackage{comment}

\usepackage{booktabs} 
\usepackage{multirow} 

\usepackage{textgreek}
\usepackage{multirow}
\usepackage{amsfonts}

\usepackage{tikz}
\usepackage{makecell}
\usepackage{longtable}
\usepackage{url}
\usepackage{xcolor}

\def\BibTeX{{\rm B\kern-.05em{\sc i\kern-.025em b}\kern-.08em
    T\kern-.1667em\lower.7ex\hbox{E}\kern-.125emX}}

\begin{document}

\title{Towards Efficient Methods in Medical Question Answering using Knowledge Graph Embeddings}

\author{
\IEEEauthorblockN{Saptarshi Sengupta\IEEEauthorrefmark{1},
Connor Heaton\IEEEauthorrefmark{1},
Suhan Cui\IEEEauthorrefmark{1},
Soumalya Sarkar\IEEEauthorrefmark{2},
Prasenjit Mitra\IEEEauthorrefmark{1}}
\IEEEauthorblockA{\IEEEauthorrefmark{1}\textit{Information Sciences and Technology} \\
\textit{Pennsylvania State University}\\
State College, USA \\
\{sks6765, czh5372, suhan, pum10\}@psu.edu}
\IEEEauthorblockA{\IEEEauthorrefmark{2}\textit{RTX Technology Research Center} \\
USA \\
soumalya.sarkar@rtx.com}
}

\maketitle

\begin{abstract}
In Natural Language Processing (NLP), Machine Reading Comprehension (MRC) is the task of answering a question based on a given context. To handle questions in the medical domain, modern language models such as BioBERT, SciBERT and even ChatGPT are trained on vast amounts of in-domain medical corpora. However, in-domain pre-training is expensive in terms of time and resources. In this paper, we propose a resource-efficient approach for injecting domain knowledge into a model without relying on such domain-specific pre-training.

Knowledge graphs are powerful resources for accessing medical information. Building on existing work, we introduce a method using Multi-Layer Perceptrons (MLPs) for aligning and integrating embeddings extracted from medical knowledge graphs with the embedding spaces of pre-trained language models (LMs). The aligned embeddings are fused with open-domain LMs BERT and RoBERTa that are fine-tuned for two MRC tasks, span detection (COVID-QA) and multiple-choice questions (PubMedQA). We compare our method to prior techniques that rely on a vocabulary overlap for embedding alignment and show how our method circumvents this requirement to deliver better performance. On both datasets, our method allows BERT/RoBERTa to either perform on par (occasionally exceeding) with stronger domain-specific models or show improvements in general over prior techniques. With the proposed approach, we signal an alternative method to in-domain pre-training to achieve domain proficiency. Our code is available here\footnote{\url{https://github.com/saptarshi059/CDQA-project}}.
\end{abstract}

\begin{IEEEkeywords}
Medical Question Answering, Knowledge Graph Embeddings, Resource Efficiency
\end{IEEEkeywords}

\section{Introduction}
Machine Reading Comprehension requires a model to answer a question ($Q$) based on a given context ($C$) where the answer ($A$) can either be a choice among a set of options (multiple-choice QA) or a span of text from the context (extractive QA). It is a challenging task because to answer a question correctly the model needs to deal with issues such as identifying entities, supporting facts in context, question intent, etc. \cite{10.1145/3357384.3358028}. With the arrival of LLMs such as ChatGPT, it was expected to bring with it strong advances in MRC performance. Unfortunately, however, while ChatGPT has shown remarkable performance on an array of tasks, recent studies \cite{yang2023exploring} suggest that even it struggles with question-answering related tasks across a range of domains which in turn leads us to fall back on BERT-style models \cite{devlin-etal-2019-bert, rogers2020primer} which are more capable for MRC \cite{luo2022choose}.



Although BERT, trained on open-domain text such as Wikipedia, is proficient in MRC \cite{devlin-etal-2019-bert}, it requires massive amounts of unlabelled medical corpora to work well on medical QA (BioBERT \cite{lee2020biobert} and SciBERT \cite{beltagy-etal-2019-scibert} adopt this strategy). While pre-training on such corpora is certainly possible for achieving strong in-domain performance, it is an expensive strategy in terms of time and hardware requirements, making it less than ideal.

A quintessential resource for accessing domain knowledge is knowledge graphs -- large, unstructured graphs of facts stored as triples (entity-relation-entity). Previous studies (\S  \ref{sec:RW}) have shown how language models can leverage knowledge graphs to improve performance on various NLP tasks. 
Knowledge graphs allow for the training of knowledge graph embeddings (KGE), which aim to capture the semantics of an entity and its behaviour with other entities, as described by the relations in the knowledge graph. \textbf{Given that knowledge graphs provide valuable information about entities, we define the main research question here as},

\begin{quote}
    \textit{Can we use Knowledge Graph Embeddings (KGE) as a cheaper alternative to pre-training corpora for injecting domain knowledge into open-domain models to match/exceed the performance of domain-specific models?}
\end{quote}

As pointed out by \cite{Liu_Zhou_Zhao_Wang_Ju_Deng_Wang_2020}, external embeddings cannot be directly fed to the input signal of transformer-based \cite{vaswani2017attention} language models because of the different objective functions used in training them. Concatenating such vectors would lead to inconsistent or \textit{heterogeneous} embedding spaces. As such, we propose a simple \textit{embedding homogenization} technique, inspired by work on feed-forward neural networks (FFNN) \cite{liu2021pay}, to fuse entity KGE into the question representation during the fine-tuning phase for MRC. 

Modern NLP models predominantly use ``wordpiece'' or Byte-Pair Encoding (BPE) schemes to construct their vocabulary. As such, their vocabulary contains a small number of ``full'' words (e.g. \textit{aviation}) and mostly subword tokens (e.g. \textit{\#\#met}) \cite{sennrich2015neural}. Seeing as knowledge graphs contain entities described using specialized language that may span multiple subwords, the requirement of vocabulary overlap is a significant hurdle. Our technique does not rely on an intersection between the set of terms to be aligned (Knowledge Graph entities/Language Model vocabulary), unlike existing approaches such as the Mikolov \cite{mikolov2013exploiting} approach. It can therefore leverage the full array of information provided by the knowledge graph, not just information for entities present in both vocabularies. 

We hypothesize that KGE alone would not be enough to see appreciable gains. Thus, we also explore the use of \textit{definition embeddings} for the identified domain terms.
Consider the domain term, \textit{Reduced} with the definition, \textit{Made less in size or amount or degree}. Such definitions are passed through our models, in feature extraction mode, and the \textit{pooler} output (\S \ref{sec:DE}) is taken as the \textit{definition embeddings}. 

We apply our method to two MRC tasks, span extraction (COVID-QA dataset \cite{moller2020covid}) and MCQ (PubMedQA \cite{jin2019pubmedqa}). Overall, our contributions are,

\begin{enumerate}
    \item We propose a domain-agnostic strategy for alignment across embeddings spaces using FFNNs that i) \textbf{does not hinge on vocabulary overlap} and ii) \textbf{can be used even if the domain terms are phrases or pseudo-words} (ex. \texttt{entry-cluster} or \texttt{related\_to}).
    
    \item Through our method, we show that open-domain models can perform either,

        \begin{itemize}
            \item Similar to domain-specific models by avoiding expensive pre-training over large in-domain corpora.
            \item Or, show improvements in performance over methods that require a vocabulary overlap for aligning external embeddings.
        \end{itemize}
    
    \item \textbf{We release a cleaned version of COVID-QA}\footnote{\url{https://huggingface.co/datasets/Saptarshi7/covid_qa_cleaned_CS}}. 
\end{enumerate}

\section{Related Work} \label{sec:RW}
When updating the input with additional information, two trends have emerged, either keep the external text as-is in raw string form (\S \ref{sec:TI}) or convert it to embeddings (\S \ref{sec:EI}) by aligning them with the model's representation space. 

\subsection{Text Integration}\label{sec:TI}

\cite{yu2021dict} injects dictionary definitions for rare words by using a custom loss function that determines if the appended definition refers to a rare or regular word. Although novel, it incurs an overhead by additional pre-training.

A more popular way of integrating plain text knowledge is via knowledge graph triples which are defined as information tuples (\texttt{<subject, predicate, object>}) described in a condensed \textit{pseudo-language} e.g. 
\texttt{(rock, heavier\_than, paper)}. \cite{bosselut2019comet} propose COMET for commonsense reasoning trained solely on knowledge graph triples. However, training a model on concepts expressed in such a pseudo-language limits generalization ability to natural text.

K-BERT \cite{Liu_Zhou_Zhao_Wang_Ju_Deng_Wang_2020} expands the identified entities in the input sentence with one-hop knowledge graph-triples and fine-tunes BERT using the updated representation and \textit{soft-position embeddings} to account for the distortion in the updated text.

\subsection{Embedding Integration}\label{sec:EI}

A different approach to knowledge augmentation is the use of \textit{external embeddings} trained on a relevant domain. \cite{poerner-etal-2020-e} propose E-BERT, a BERT model fine-tuned using external embeddings for question-answering. They homogenize entity embeddings using the Mikolov cross-lingual alignment strategy \cite{mikolov2013exploiting} which requires a substantial overlap between the entities in the external knowledge source and the LM's vocabulary. In domains such as ours where we do not see a huge overlap, such an alignment technique is restrictive.

\cite{sharma-etal-2019-incorporating} utilize KGE for medical inference. 
After identifying all the biomedical concepts in the dataset, they create a sub-graph from the UMLS knowledge graph (Unified Medical Language System) \cite{bodenreider2004unified} on which they train KGE's. These embeddings are then concatenated with BioELMo \cite{jin2019probing} embeddings and used with the ESIM model \cite{chen2017enhanced}. Although they directly concatenated external embeddings, they train the entire ESIM model to project these vectors in the same space thereby avoiding heterogeneous embeddings.

\subsection{Embedding Alignment}
\label{sec:EA}

Mikolov \cite{mikolov2013exploiting} propose a strategy for cross-lingual embedding alignment for language translation, using a linear objective function presented in eq. \eqref{eq:mik}, where $\{ (x_i, z_i) $\} are a set of $n$ translation pairs - $x_i$, the \textit{source} and $z_i$ the \textit{target} embedding. Once $W$ is learned, the target embedding $z_i$ for a given source embedding $x_i$ is computed simply as $z_i = Wx_i$. Further improvements were made to this method by \cite{xing2015normalized} and \cite{zhang2019girls} by enforcing various normalization constraints.


\begin{equation}
\label{eq:mik}
   \min_{W} \sum_{i=1}^{n} \lVert {Wx_i} - z_i \rVert^2 
\end{equation}



Although these modifications address different shortcomings of the base linear transformation strategy, they still depend on two key criteria: 1) a mapping dictionary and 2) a common dimensionality between \textit{source} and \textit{target} embeddings. The first criterion is more difficult to satisfy today than when the Mikolov approach was first introduced. The vocabulary of modern transformers does not consist of complete words, but rather word \textit{tokens} that often do not constitute a word on their own. This poses a challenge when trying to integrate embeddings from a knowledge base, for example, where embeddings correspond to whole words or even word phrases. This is compounded by the second criterion which requires the embeddings to be of the same dimension. When working with embeddings from two transformer-based LMs, this is often the case, but less common when working with embeddings from other resources. Such criteria significantly limit the utility of the previous approaches.

We use the original Mikolov strategy as a baseline for two reasons, i) it avoids the similar dimensionality constraint and ii) it was used by a related method (E-BERT) and can thus be used for comparative analysis.

\section{Proposed Methodology}

The overall pipeline can be broken down into four phases, 1) \textbf{Entity linking} 2) \textbf{KGE homogenization} 3) \textbf{Definition embedding generation} and 4) \textbf{Fine-Tuning with external knowledge infusion}. 

\subsection{Resources Used}

The following resources are used in our study, 


\begin{enumerate}
    \item \textbf{COVID-QA}: As mentioned previously, we perform span detection on COVID-QA, a SQuAD \cite{rajpurkar-etal-2018-know} style dataset with 2,019 question-answer pairs based on 147 scientific articles and annotated by 15 biomedical experts. 
    
    \item \textbf{PubMedQA}: Multiple-choice QA was performed on the PubMedQA benchmark that has a collection of 1k expert-annotated instances of \textit{yes/no/maybe} biomedical questions. The 1k samples are split into 500 samples each of training and test with the training data being further broken up into 10-folds with 450 samples in training and 50 in the development set for each fold. 
    
    \item \textbf{UMLS}: We use the metathesaurus (a collection of various biomedical terminologies) from the UMLS to extract entity definitions. our choice for using the UMLS was twofold; I) a well-maintained entity linker (MetaMap) is available for it, and II) UMLS has been developed and updated for a long time and is thus robust in terms of coverage and accuracy.

    
    
    \item \textbf{Pre-Trained UMLS KGE}: \cite{maldonado2019adversarial} train 3.2M $R^{50}$ entity embeddings using knowledge graph triples from the UMLS metathesaurus and semantic network \footnote{\url{https://lhncbc.nlm.nih.gov/semanticnetwork/}}.
    

    \item \textbf{MetaMap} Following \cite{sharma-etal-2019-incorporating}, we use MetaMap \cite{aronson2010overview} as our entity identifier/linker. MetaMap works in tandem with UMLS since it breaks down input sentences according to the UMLS entities it discovers within it. 

\end{enumerate}

\subsection{Preprocessing (COVID-QA Cleanup)}

In its original state, COVID-QA was rife with syntactical and encoding issues. We focus mainly on cleaning the questions since the contexts were research papers that have already been screened. 
We use a dedicated grammar checking service (Grammarly) to identify a total of 1020 questions (50.5\%) that had the following issues,

\begin{itemize}
    \item Excess spaces (\textit{For what sca algorithm was applied to improve the anfis model\textless space\textgreater?})
    \item Missing spaces (\textit{What percentage of patients do not return for \textbf{followup} after hiv testing?})
    \item Uncapitalized acronyms (\textit{What is \textbf{ifitm}?})
    \item Repeated words (\textit{\textbf{Was was} the sample size?})
    \item Spelling mistakes (\textit{How does \textbf{mannanose} binding lectin (mbl) affect elimination of hiv-1 pathogen?})
    \item Grammatical issues (\textit{What suggests that ip-10 plays a significant role \textbf{on} the pathogenesis of pneumonia?})
\end{itemize}

Finally, we ran the National Laboratory of Medicine's \texttt{replace\_UTF8} \footnote{\url{https://lhncbc.nlm.nih.gov/ii/tools/MetaMap/additional-tools/ReplaceUTF8.html}} tool to replace Unicode characters resulting in terms such as \textbf{$\beta$-amyloid} becoming \textbf{beta-amyloid}. Table \ref{tab:clevuc} demonstrates the effect of this pre-processing. 


\begin{table*}[ht]
\centering
\caption{Baseline (no fine-tuning/zero-shot) performance change due to COVID-QA cleanup}
\begin{tabular}{@{}lcccccc@{}}
\toprule
\textbf{Model} & \multicolumn{2}{c}{\textbf{Uncleaned Dataset}} & \multicolumn{2}{c}{\textbf{Cleaned Dataset}} & \multicolumn{2}{c}{\textbf{Improvement (\%)}} \\
\cmidrule(lr){2-3} \cmidrule(lr){4-5} \cmidrule(lr){6-7}
& \multicolumn{1}{c}{\textbf{F1}} & \multicolumn{1}{c}{\textbf{EM}} & \multicolumn{1}{c}{\textbf{F1}} & \multicolumn{1}{c}{\textbf{EM}} & \multicolumn{1}{c}{\textbf{F1}} & \multicolumn{1}{c}{\textbf{EM}} \\
\midrule
BERT\textsubscript{BASE} & 0.40 & 0.21 & 0.40 & 0.22 & 0.0 & \textbf{4.8} \\
RoBERTa\textsubscript{BASE} & 0.44 & 0.24 & 0.44 & 0.24 & 0.0 & 0.0 \\
BioBERT & 0.40 & 0.22 & 0.43 & 0.24 & \textbf{7.5} & \textbf{9.1} \\
SciBERT & 0.44 & 0.25 & 0.45 & 0.25 & \textbf{2.3} & 0.0 \\
\bottomrule
\end{tabular}
\label{tab:clevuc}
\end{table*}

\subsection{Entity Linking} 

We ran MetaMap on the 2,019 COVID-QA and 1k PubMedQA questions revealing 1,897 and 2,782 entities resp. Among them, 1,837 and 2,694 were common with the pre-trained KGEs. Only 1,452 and 2,078 entities had definitions in the metathesaurus and were chosen for homogenization. MetaMap had the following settings: suppress all numeric concepts, unique acronyms only (retain the most relevant acronym in case of multiple matches based on the context) and no derivational variants (consider only exact matches with UMLS concepts and not derived word forms). These settings led to the least noise in the linking process and ultimately best performance for our models. We also use a custom \footnote{released alongside code} acronym and concept exclusion list, which was curated after studying the vanilla linker outputs. Numbers were excluded since it is known that BERT does not deal with numbers well \cite{wallace2019nlp}.

\subsection{KGE Homogenization}

According to the universal approximation theorem \cite{hornik1989multilayer}, an MLP with sigmoid activation and at least one hidden layer of arbitrary length can approximate any well-behaved function. Using this as a guiding principle, we propose a method for learning homogenized $\mathbb{R}^{d^{LM}}$ vectors from $\mathbb{R}^{d^{KGE}}$ ones, where $d^{LM} = 768$ and $d^{KGE} = 50$.

The FFNN used to homogenize the KGE consisted of only \textit{one hidden layer}. Inputs to the network were first subjected to a \textit{dropout} regularization, with a probability of $0.25$, following which they passed through the hidden layer ($d^{hidden} = 300$). Outputs from the hidden layer were processed via a \textit{layer normalization} and \textbf{TanH} activation. A final linear transformation was applied to the outputs to produce a $\mathbb{R}^{d^{LM}}$ vector. 
The network was trained with a batch size of $256$ for $30$ epochs and optimized using Adam with an L2 penalty of $0.001$. The network minimizes the \textbf{MSE} loss between the FFNN output and the \textit{average of the entity subword embeddings}, obtained from the model vocabulary. Consider the entity \textit{cysteine}. When tokenized using BERT, it generates the subwords, [\textit{`cy', `\#\#stein', `\#\#e'}]. An average of these vocabulary embeddings represents the \textit{target} output for the entity.

\begin{figure}
    \centering
    \includegraphics[scale=0.52]{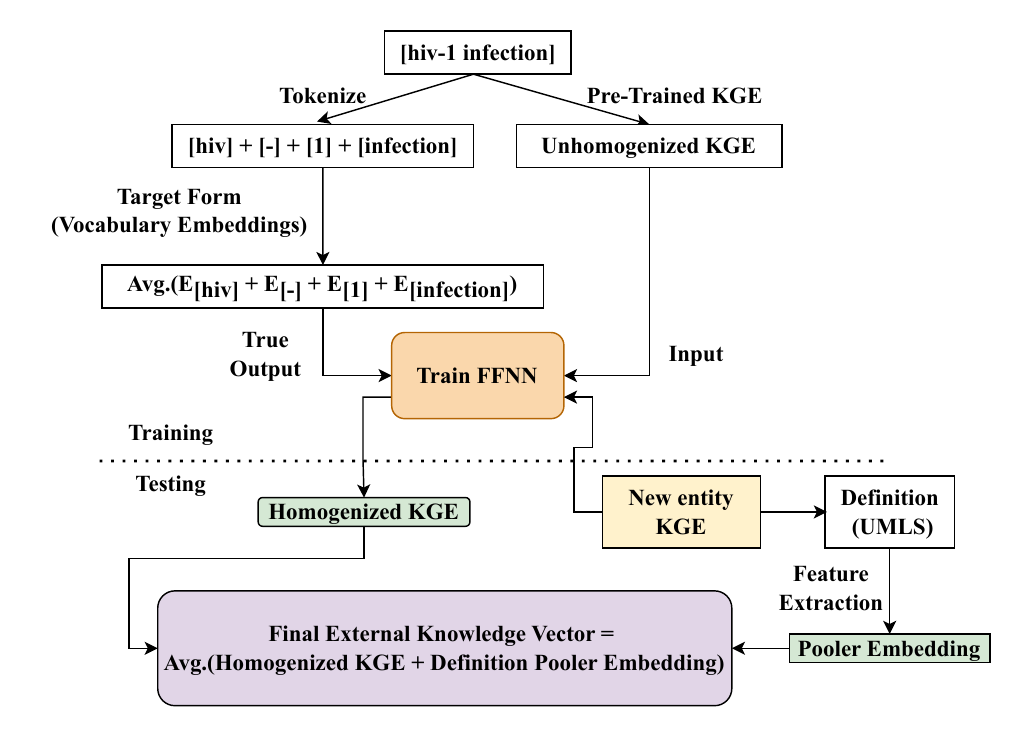}
    \caption{Proposed Homogenization Method explained using an example entity \texttt{hiv-1 infection}. Here, \texttt{E} stands for the models' Vocabulary Embedding.}
    \label{fig:homogenization_diagram}
\end{figure}

After exploring various training strategies, we found that the above settings yielded the best test loss on the desired entity embeddings. We selected 10K samples from the pre-trained KGE, which had no overlap with the COVID-QA (1,452) and PubMedQA (2,078) dataset entities, to train our projection network. Once the network is trained, the final linear transformation stands as the homogenized embeddings for our KGEs. The overall process is described in Figure \ref{fig:homogenization_diagram}.

\subsection{Definition Embeddings} \label{sec:DE}

We hypothesized that KGEs alone would not be enough to see significant performance gains on the task. Thus, we decided to incorporate \textit{entity definitions} to provide an added source of external knowledge. However, simply using the text form of the definitions would create longer questions and shorter context representations, which in turn would generate more negative samples since BERT(s) can only handle $512$ input tokens. 

We vectorized the definitions by passing them through the respective LMs, in a \textit{feature extraction mode}, and using the model-specific \textit{pooler output}, which for BERT is the $[CLS]$ (classification) token further processed by a linear layer (weights obtained from the NSP task) and TanH activation. We usually observed that it is more beneficial to use the processed $[CLS]$ embedding i.e. the pooler output rather than the regular $[CLS]$ and thus decided to use the former.

\subsection{Fine-Tuning with Improved Question Representation}

For each entity, we have two embeddings, a homogenized KGE and a \textit{definition embedding}. We observed increased training time and subpar results when we added the embeddings \textit{separately} into the input representation, by the schemes described below. As such, we decided to \textit{average} the two embeddings to form the final external knowledge vector. Two \textit{existing} schemes of integrating embeddings were explored. Both schemes are explained using the sample question, \textit{What is the main cause of HIV-1 infection in children?}, which when tokenized using BERT's tokenizer yields the following set of tokens, \textit{[`what', `is', `the', `main', `cause', `of', `hiv', `-', `1', `infection', `in', `children', `?']}. We fine-tune regularly, i.e. we do not make architectural changes to the model.

\begin{enumerate}
    \item \textbf{BERTRAM concatenation}: According to \cite{schick2020bertram}, the external embedding for a given token should be concatenated \textit{alongside} it using the \textit{/} (slash) symbol as a separator as follows,

\begin{quote}
    \color{black} [CLS] \color{blue} [what] [is] [the] \color{violet} [main] \color{blue} [/] \color{red} [main]  \color{violet} [cause] \color{blue} [/] \color{red} [cause] \color{blue} [of] \color{violet} [hiv 1 infection] \color{blue} [/] \color{red} [hiv] [1] [infection] \color{blue} [in] \color{violet} [children] \color{blue} [/] \color{red} [children] \color{blue} [?] \color{black} [SEP] \color{teal} [context tokens]
\end{quote}

where the terms in \textcolor{blue}{blue} are the \textcolor{blue}{non-entity terms}, terms in \textcolor{red}{red} are the identified \textcolor{red}{entities} and the ones in \textcolor{violet}{violet} are their corresponding \textcolor{violet}{external knowledge embeddings}. 

\item \textbf{DEKCOR concatenation}: \cite{xu2020fusing} concatenated external embeddings \textit{without tampering the original text}. Such embeddings are added alongside the tokenized input and separated using the special \textit{[SEP]} token as follows,

\begin{quote}
    \color{black} [CLS] \color{blue} [what] [is] [the] \color{red}[main] [cause] \color{blue} [of] \color{red} [hiv] [-] [1] [infection] \color{blue} [in] \color{red} [children] \color{blue} [?] \color{black} [SEP] \color{violet} [main] [cause] [hiv 1 infection] [children] \color{black} [SEP] \color{teal} [context tokens]
\end{quote}
\end{enumerate}

\section{Results}

Table \ref{tab:covidqa_results} and \ref{tab:pubmedqa_results} show the results from our COVID-QA and PubMedQA experiments respectively. 
Experimentation begins with models that are fine-tuned on a related task (SQuAD w.r.t COVID-QA and SNLI w.r.t PubMedQA) as it has been shown that models benefit from such a strategy \cite{castelli-etal-2020-techqa,soni2022radqa}. While SQuAD-trained models provide perfect pairing for COVID-QA, we could not find a related model for PubMedQA since classification tasks seldom have the same label space. Thus, while we could provide zero-shot scores for COVID-QA, we could not for PubMedQA since models fine-tuned on SNLI had learned a different label space.

We choose BERT\textsubscript{BASE} and RoBERTa\textsubscript{BASE} as our general-purpose models and Bio/Sci-BERT for our domain-specific models. As mentioned above, the models were first trained on the SQuAD and SNLI benchmarks before being fine-tuned for COVID-QA and PubMedQA and were trained using 5-fold and 10-fold cross-validation (CV) for each dataset respectively. We report the average F1 and EM for COVID-QA and the average accuracy and F1 for PubMedQA across all folds.

In addition to the Mikolov (E-BERT) baseline, we also conduct trials in which the external embeddings were randomized before fine-tuning. We believed that such a setup would allow us to gauge how much attention the model is dedicating towards the additional knowledge signals. The intuition was that if the model performance completely crashed using random embeddings, well-homogenized vectors would certainly then be of benefit. On the other hand, if model performance remained relatively the same w.r.t vanilla fine-tuning or our approach, we could take it to mean that adding extra embeddings, that were not seen during the pre-training process, provides no extra advantage.

Finally, regarding concatenation strategies, we observe an overall improvement using DEKCOR concatenation for COVID-QA. We reason that this is due to the query retaining its original representation instead of being intertwined with the external embeddings. 
However, as scores from our PubMedQA trials indicate gains using either approach, we reason that there is merit to each strategy and choosing one over the other depends on the task and model at hand. 

\begin{table*}[ht]
    \centering
    \caption{Average F1/EM over 5-fold CV on COVID-QA of various model architectures with varying external resources and integration methods.}
    \begin{tabular}{|c|c|c|c|c|c|c|c|}
    \hline
    \thead{\textbf{Architecture}} & \thead{\textbf{Fine-Tuning}} & \thead{\textbf{UMLS} \\ \textbf{Embeds (KGE)}} & \thead{\textbf{Definition} \\ \textbf{Embeds}} & \thead{\textbf{BERTRAM} \\ \textbf{Concat}} & \thead{\textbf{DEKCOR} \\ \textbf{Concat}} & \thead{\textbf{F1}} & \thead{\textbf{EM}}  \\
    \hline
    \multirow{11}{*}{BERT\textsubscript{BASE}} & &  &  &  &  & 0.402 & 0.219 \\ 
    \cline{2-8}
     & \checkmark &  &  &  &  & 0.476 & 0.235 \\ 
    \cline{2-8}
     & \checkmark & Random & & \checkmark &  & 0.469 & 0.240 \\ 
    \cline{2-8}
     & \checkmark & Random & & & \checkmark & 0.478 & 0.244 \\ 
    \cline{2-8}
     & \checkmark & E-BERT & & \checkmark & & 0.471 & 0.247 \\ 
     \cline{2-8}
     & \checkmark & E-BERT & & & \checkmark & 0.482 & 0.249 \\ 
    \cline{2-8}
     & \checkmark & \checkmark & & \checkmark & & 0.461 & 0.237 \\ 
    \cline{2-8}
     & \checkmark & \checkmark & & & \checkmark & 0.483 & 0.246 \\ 
     \cline{2-8}
      & \checkmark & & \checkmark & \checkmark & & 0.467 & 0.238 \\ 
    \cline{2-8}
     & \checkmark & & \checkmark & & \checkmark & 0.484 & 0.249 \\ 
    \cline{2-8}
     & \checkmark & \checkmark & \checkmark & \checkmark & & 0.473 & 0.243 \\ 
    \cline{2-8}
     & \checkmark & \checkmark & \checkmark & & \checkmark & \textbf{0.485} & \textbf{0.250} \\ 
    
    \hline
    \hline
    
    \multirow{11}{*}{RoBERTa\textsubscript{BASE}} & & & & & & 0.437 & 0.243 \\ 
    \cline{2-8}
     & \checkmark & & & & & \textbf{0.529} & 0.278 \\ 
    \cline{2-8}
     & \checkmark  & Random & & \checkmark &  & 0.503 & 0.275 \\ 
    \cline{2-8}
     & \checkmark  & Random & & & \checkmark & 0.515 & 0.281 \\ 
    \cline{2-8}
     & \checkmark & E-BERT & & \checkmark & & 0.501 & 0.275 \\ 
    \cline{2-8}
     & \checkmark & E-BERT & & & \checkmark & 0.523 & 0.290 \\ 
    \cline{2-8}
     & \checkmark & \checkmark & & \checkmark & & 0.518 & 0.287 \\ 
    \cline{2-8}
     & \checkmark & \checkmark & & & \checkmark & 0.514 & 0.276 \\ 
    \cline{2-8}
     & \checkmark & & \checkmark & \checkmark & & 0.511 & 0.277 \\ 
    \cline{2-8}
     & \checkmark & & \checkmark & & \checkmark & 0.520 & 0.286 \\ 
    \cline{2-8}
     & \checkmark & \checkmark & \checkmark & \checkmark & & 0.498 & 0.273 \\ 
    \cline{2-8}
     & \checkmark & \checkmark & \checkmark & & \checkmark & \textbf{0.529} & \textbf{0.299} \\ 

    \hline
    \hline
    \multirow{4}{*}{BioBERT} &  &  &  & &  & 0.427 & 0.238 \\ 
    \cline{2-8}
     & \checkmark  &  &  & &  & 0.504 & 0.275 \\ 
    \cline{2-8}
     & \checkmark  & \checkmark & \checkmark  & \checkmark &  & 0.499 & 0.269 \\ 
    \cline{2-8}
     & \checkmark  & \checkmark & \checkmark  & & \checkmark & \textbf{0.509} & \textbf{0.276} \\ 
    \hline
    \hline
    \multirow{4}{*}{SciBERT} &  &  &  & &  & 0.450 & 0.248 \\
    \cline{2-8}
     & \checkmark  &  &  & &  & \textbf{0.537} & 0.286 \\ 
    \cline{2-8}
     & \checkmark  & \checkmark & \checkmark  & \checkmark & & 0.532 & \textbf{0.290} \\ 
    \cline{2-8}
     & \checkmark  & \checkmark & \checkmark  & & \checkmark & 0.531 & 0.282 \\ 
    \hline
    \end{tabular}
    \label{tab:covidqa_results}
\end{table*}

\begin{table*}[ht]
    \centering
    \caption{Average Accuracy/F1 over 10-fold CV on the test set of PubMedQA, for a given epoch \& across all folds, with varying external resources and integration methods.}
    \begin{tabular}{|c|c|c|c|c|c|c|c|}
    \hline
    \thead{\textbf{Architecture}} & \thead{\textbf{Fine-Tuning}} & \thead{\textbf{UMLS} \\ \textbf{Embeds (KGE)}} & \thead{\textbf{Definition} \\ \textbf{Embeds}} & \thead{\textbf{BERTRAM} \\ \textbf{Concat}} & \thead{\textbf{DEKCOR} \\ \textbf{Concat}} & \thead{\textbf{Accuracy}} & \thead{\textbf{F1}}  \\
    \hline
    \multirow{11}{*}{BERT\textsubscript{BASE}}
     & \checkmark &  &  &  &  & 51.28 & 0.30 \\ 
    \cline{2-8}
     & \checkmark & Random & & \checkmark &  & 53.66 & 0.32 \\ 
    \cline{2-8}
     & \checkmark & Random & & & \checkmark & 53.72 & 0.32 \\ 
    \cline{2-8}
     & \checkmark & E-BERT & & \checkmark & & 53.66 & \textbf{0.33} \\ 
     \cline{2-8}
     & \checkmark & E-BERT & & & \checkmark & 53.78 & 0.32 \\ 
    \cline{2-8}
     & \checkmark & \checkmark & & \checkmark & & 53.68 & 0.32 \\ 
    \cline{2-8}
     & \checkmark & \checkmark & & & \checkmark & 53.90 & 0.32 \\
     \cline{2-8}
      & \checkmark & & \checkmark & \checkmark & & 53.50 & \textbf{0.33} \\ 
    \cline{2-8}
     & \checkmark & & \checkmark & & \checkmark & 53.74 & 0.32 \\ 
    \cline{2-8}
     & \checkmark & \checkmark & \checkmark & \checkmark & & \textbf{53.92} & 0.32 \\ 
    \cline{2-8}
     & \checkmark & \checkmark & \checkmark & & \checkmark & 52.98 & 0.32 \\ 
    
    \hline
    \hline
    
    \multirow{11}{*}{RoBERTa\textsubscript{BASE}}
     & \checkmark & & & & & 52.90 & 0.27 \\ 
    \cline{2-8}
     & \checkmark & Random & & \checkmark &  & 52.40 & 0.32 \\ 
    \cline{2-8}
     & \checkmark  & Random & & & \checkmark & 52.64 & 0.29 \\ 
    \cline{2-8}
     & \checkmark & E-BERT & & \checkmark & & 50.46 & \textbf{0.33} \\ 
    \cline{2-8}
     & \checkmark & E-BERT & & & \checkmark & 51.12 & 0.32 \\ 
    \cline{2-8}
     & \checkmark & \checkmark & & \checkmark & & \textbf{53.74} & 0.31 \\ 
    \cline{2-8}
     & \checkmark & \checkmark & & & \checkmark & 52.92 & 0.29 \\ 
    \cline{2-8}
     & \checkmark & & \checkmark & \checkmark & & 51.04 & 0.32 \\ 
    \cline{2-8}
     & \checkmark & & \checkmark & & \checkmark & 53.24 & 0.30 \\
    \cline{2-8}
     & \checkmark & \checkmark & \checkmark & \checkmark & & 51.88 & 0.32 \\ 
    \cline{2-8}
     & \checkmark & \checkmark & \checkmark & & \checkmark & 52.36 & 0.31 \\

    \hline
    \hline
    \multirow{3}{*}{BioBERT} 
     & \checkmark  &  &  & &  & \textbf{65.06} & \textbf{0.45}  \\
    \cline{2-8}
     & \checkmark  & \checkmark & \checkmark  & \checkmark &  & 64.76 & 0.44 \\ 
    \cline{2-8}
     & \checkmark  & \checkmark & \checkmark  & & \checkmark & 64.42 & 0.44  \\ 
    \hline
    \hline
    \multirow{3}{*}{SciBERT}
     & \checkmark  &  &  & &  & 63.66 & \textbf{0.44} \\ 
    \cline{2-8}
     & \checkmark  & \checkmark & \checkmark  & \checkmark & & \textbf{63.86} & \textbf{0.44} \\ 
    \cline{2-8}
     & \checkmark  & \checkmark & \checkmark  & & \checkmark & 63.74 & \textbf{0.44} \\ 
    \hline
    \end{tabular}
    \label{tab:pubmedqa_results}
\end{table*}

\subsection{COVID-QA Results}

Here, utilizing both KGE and \textit{definition embedding}, integrated via DEKCOR concatenation led to the best performance for the non-domain-specific variants. This approach improves BERT's F1 and EM scores by 1.9\% and 6.4\% respectively over regular fine-tuning, while RoBERTa sees no performance gains in terms of F1, but a 7.6\% improvement in EM score w.r.t regular fine-tuning.

Comparing our scores for the non-domain-specific models against the E-BERT (Mikolov strategy) baselines we see that BERT achieves 0.6\% and 0.4\% improvement whereas RoBERTa achieves 1.2\% and 3.1\% improvement in F1 and EM scores respectively. Despite BERT's vocabulary having only 24 terms in common with our KGE and RoBERTa having 201 terms in common, they still perform relatively well using the Mikolov strategy. 

Concerning the domain-specific models, BioBERT shows a 1\% and 0.36\% improvement in F1 and EM resp. over vanilla fine-tuning whereas SciBERT does not show an increase in F1 but a 1.4\% improvement in EM over vanilla fine-tuning. Interestingly, the improved EM score for SciBERT arises from regular concatenation.

Overall, we see that by incorporating both KGE and DE, BERT's performance comes very close to BioBERT's and RoBERTa's parallels SciBERT's. \textbf{In fact, RoBERTa's EM is around 3.1\% more than SciBERT's. These results show that open-domain models can perform at the level of domain-specific models through external knowledge fusion}.

\subsection{PubMedQA Results}

While results from COVID-QA were found to be more uniform w.r.t concatenation strategy and type of embeddings added - i.e. [DEKCOR + avg.(KGE + \textit{definition embedding})] = best performance - a similar observation was not seen for PubMedQA. For BERT, we see that KGE and \textit{definition embedding} along with BERTRAM concatenation led to the best accuracy across all model configurations while the F1 was best for E-BERT and \textit{definition embedding} + BERTRAM concatenation. Over regular fine-tuning, BERT's accuracy improves by 5.2\% and F1 by 10\%, for the aforementioned configurations. For RoBERTa, we did not see any performance gains coming from our proposed method. However, in isolation, BERTRAM concatenation yielded the best accuracy (1.6\% over regular fine-tuning) while E-BERT held the best score in terms of F1 (22.2\% above regular fine-tuning). 

Compared to the best E-BERT baseline, BERT sees a 0.26\% improvement in accuracy while the F1 stays the same. \textbf{Our best-performing RoBERTa model sees a 5.1\% increase in accuracy over E-BERT which shows the effectiveness of our method in utilizing the entire vocabulary of the model instead of only a subset of it.}. 

For this task, we see that our external knowledge vectors aren't able to raise the performance of BERT/RoBERTa to domain-specific Bio/Sci-BERT level. We see that BioBERT on its own performs better than with our modifications, whereas SciBERT improves in accuracy over regular fine-tuning (0.3\%) while the F1 remains on par with the original.

\subsection{Ablation Studies}

Following \cite{poerner-etal-2020-e}, we realize that replacing entity tokens with their externally homogenized form would result in poor performance since the semantics of the sentence would get drastically altered rather than enhanced. As such, we only perform concatenation experiments.

We investigate \textit{definition embeddings} and KGE in isolation to understand the influence of each. For COVID-QA, w.r.t vanilla fine-tuning of BERT, KGE alone improve F1 and EM performance by 1.5\% and 4.7\% resp. whereas \textit{definition embeddings} improves it by 1.7\% and 6\% resp. On the other hand, over vanilla fine-tuning for RoBERTa, KGE alone brings down F1 by 2.1\% but improves EM by 3.2\% and \textit{definition embeddings} alone decreases F1 by 1.7\% but increases EM by 2.9\%. Thus, we see that for BERT, definition embeddings seem to have more impact than the homogenized KGE while for RoBERTa the situation is split since the KGE improve EM the most while the \textit{definition embedding} hampers F1 the least. We conjecture that this benefit from \textit{definition embedding} is due to them resembling transformer vectors the most. The homogenized KGE while providing some benefit, still has to go through a transformation which, we hypothesize, adds a degree of noise.

For PubMedQA, we see that the best KGE model yields a 5.1\% and 6.7\% increase in accuracy and F1 resp. over regular fine-tuning. With just \textit{definition embeddings} we observe a similar improvement in F1 while accuracy improves by 4.8\%. For RoBERTa, accuracy and F1 improve over regular fine-tuning by 1.6\% and 14.8\% resp. using only KGE while \textit{definition embeddings} alone from the best model obtains 0.6\% and 11.1\% improvements resp.

\section{Discussion}
An interesting takeaway from this study is the performance difference from integrating external knowledge with BERTRAM v/s DEKCOR concatenation. This disparity is more pronounced for RoBERTa than BERT-based methods, which we attribute to the tokenization scheme employed by each model. RoBERTa's vocabulary, constructed using BPE, includes spaces. Thus, interjecting new text within the question fundamentally alters how the language is decomposed, and which tokens are presented to the model. In comparison, BERT's tokens do not span spaces and is not as impacted when new tokens are inserted. Using the running question, we have the following tokenization for RoBERTa, \textit{[`What', `Ġis', `Ġthe', `Ġmain', `Ġcause', `Ġof', `ĠHIV', `-', `1', `Ġinfection', `Ġin', `Ġchildren', `?']} where \textit{Ġ} is a special character indicating a space between tokens.

We attribute the performance disparity in RoBERTa to the corruption of the input that takes place when external knowledge tokens are included. The text is no longer ``well-formed'' natural language. Additionally, for COVID-QA, we see the most performance improvement in terms of EM. We hypothesize that this is because the information provided by the external embeddings helps pinpoint the domain-specific answers. PubMedQA scores, on the other hand, see overall enhancements for both metrics (F1 being greater) for BERT and RoBERTa based models. This makes sense since by fine-tuning, models start understanding class distribution better which in turn leads to improved F1 scores.

Generally, integrating external embeddings provides performance improvements for non-domain-specific models. Somewhat peculiarly, we see that even adding \textit{random} entity embeddings improves performance over vanilla fine-tuning of domain-specific models. We hypothesize that although the random embeddings do not convey any real information, they can serve to denote the presence of different, relevant domain terms. It then holds that the Mikolov approach cannot denote all relevant terms as it requires a vocabulary overlap.

Under the proposed framework, homogenization does not need to rely on a vocabulary overlap between the embedding spaces to be aligned. Thus, it allows the method to scale well to domains where there is no significant vocabulary overlap, such as ours. Ultimately, we see that by integrating external knowledge, open-domain LMs can either perform at a level comparable to medical LMs or show improved performance, in general, on biomedical tasks.

\subsection{Complexity Analysis}

We analyze the computational complexity of our approach against the Mikolov baseline. According to eq. \eqref{eq:mik}, the converted embedding ($\hat{y}$) for the entity embedding to be homogenized ($x^{entity}$) is given by, 

\begin{equation}
    \hat{y} = f(x^{entity}; W)
\end{equation}

Whereas, our FFNN has,

\begin{equation}
    \hat{y} = W^{(2)}[\tanh(W^{(1)}x^{entity} + b_1)] + b_2    
\end{equation}

where $W^{(1)}$ and $W^{(2)}$ are the weight matrices of the linear layers and $b_1$, $b_2$ are their respective bias terms. In other words, $\hat{y} = f(x^{entity}; W^{(1)}, W^{(2)})$. This indicates that training the NN, incurs an additional computational overhead since we are trying to learn two parameters instead of one. However, such a cost is mitigated by the fact that most modern-day machines are more than capable of optimizing a simple NN, such as this, without much energy overhead.

Finally, since the homogenization is \textit{specific} to the model with which we want to align our embeddings, we must retrain the network for each model we want to align with. This factor \textit{might} become a bottleneck if there are a large number of entity embeddings to be homogenized. However, even then, we envisage that the computational cost \textit{would not} be a concern for the reason mentioned previously which in turn shows the scalability of our method.

\section{Conclusions and Future Work}

In this paper, we investigate the use of external knowledge embedding integration for the task of MRC. Results indicate a degree of promise in the proposed approach, demonstrating how domain-specific information can be added to the input representation of a non-domain-specific model, avoiding the lengthy pre-training process. However, there is much work yet to be done. First, we believe that more research is needed into figuring out alternative strategies for incorporating external embeddings, in the \textit{input representation}. While it might be computationally feasible to train adapter layers, they do present a processing overhead and increase the overall complexity of the base model. 


While models such as Bio/Sci-BERT were trained to obtain SOTA performance on domain-specific-tasks, we see a wide gap between their performance on COVID-QA v/s PubMedQA. Higher scores on the latter indicate the benefit of pre-training on domain-specific corpora. However, results on the former makes it difficult to consider this a definitive claim. Thus, more inspection is needed into the effect of pre-training on such data.

Finally, further investigation into the overall poor performance of these models is needed. 
A quick study of the datasets revealed that while the subject matter is dense, the questions and their corresponding answers are straightforward. Thus, investigating issues in the underlying architecture or semantic disconnect between medical and open-domain corpora will pave the way for true domain generalization.

\section*{Acknowledgment}

This work is supported by RTX Technology Research Center.

\bibliographystyle{IEEEtran}
\bibliography{IEEEabrv,IEEEexample}

\begin{thebibliography}{10}
\providecommand{\url}[1]{#1}
\csname url@samestyle\endcsname
\providecommand{\newblock}{\relax}
\providecommand{\bibinfo}[2]{#2}
\providecommand{\BIBentrySTDinterwordspacing}{\spaceskip=0pt\relax}
\providecommand{\BIBentryALTinterwordstretchfactor}{4}
\providecommand{\BIBentryALTinterwordspacing}{\spaceskip=\fontdimen2\font plus
\BIBentryALTinterwordstretchfactor\fontdimen3\font minus \fontdimen4\font\relax}
\providecommand{\BIBforeignlanguage}[2]{{%
\expandafter\ifx\csname l@#1\endcsname\relax
\typeout{** WARNING: IEEEtran.bst: No hyphenation pattern has been}%
\typeout{** loaded for the language `#1'. Using the pattern for}%
\typeout{** the default language instead.}%
\else
\language=\csname l@#1\endcsname
\fi
#2}}
\providecommand{\BIBdecl}{\relax}
\BIBdecl

\bibitem{10.1145/3357384.3358028}
\BIBentryALTinterwordspacing
B.~van Aken, B.~Winter, A.~L\"{o}ser, and F.~A. Gers, ``How does bert answer questions? a layer-wise analysis of transformer representations,'' in \emph{Proceedings of the 28th ACM International Conference on Information and Knowledge Management}, ser. CIKM '19.\hskip 1em plus 0.5em minus 0.4em\relax New York, NY, USA: Association for Computing Machinery, 2019, p. 1823–1832. [Online]. Available: \url{https://doi.org/10.1145/3357384.3358028}
\BIBentrySTDinterwordspacing

\bibitem{yang2023exploring}
X.~Yang, Y.~Li, X.~Zhang, H.~Chen, and W.~Cheng, ``Exploring the limits of chatgpt for query or aspect-based text summarization,'' \emph{arXiv preprint arXiv:2302.08081}, 2023.

\bibitem{devlin-etal-2019-bert}
\BIBentryALTinterwordspacing
J.~Devlin, M.-W. Chang, K.~Lee, and K.~Toutanova, ``{BERT}: Pre-training of deep bidirectional transformers for language understanding,'' in \emph{Proceedings of the 2019 Conference of the North {A}merican Chapter of the Association for Computational Linguistics: Human Language Technologies, Volume 1 (Long and Short Papers)}.\hskip 1em plus 0.5em minus 0.4em\relax Minneapolis, Minnesota: Association for Computational Linguistics, Jun. 2019, pp. 4171--4186. [Online]. Available: \url{https://aclanthology.org/N19-1423}
\BIBentrySTDinterwordspacing

\bibitem{rogers2020primer}
A.~Rogers, O.~Kovaleva, and A.~Rumshisky, ``A primer in bertology: What we know about how bert works,'' \emph{Transactions of the Association for Computational Linguistics}, vol.~8, pp. 842--866, 2020.

\bibitem{luo2022choose}
M.~Luo, K.~Hashimoto, S.~Yavuz, Z.~Liu, C.~Baral, and Y.~Zhou, ``Choose your qa model wisely: A systematic study of generative and extractive readers for question answering,'' in \emph{Proceedings of the 1st Workshop on Semiparametric Methods in NLP: Decoupling Logic from Knowledge}, 2022, pp. 7--22.

\bibitem{lee2020biobert}
J.~Lee, W.~Yoon, S.~Kim, D.~Kim, S.~Kim, C.~H. So, and J.~Kang, ``Biobert: a pre-trained biomedical language representation model for biomedical text mining,'' \emph{Bioinformatics}, vol.~36, no.~4, pp. 1234--1240, 2020.

\bibitem{beltagy-etal-2019-scibert}
\BIBentryALTinterwordspacing
I.~Beltagy, K.~Lo, and A.~Cohan, ``{S}ci{BERT}: A pretrained language model for scientific text,'' in \emph{Proceedings of the 2019 Conference on Empirical Methods in Natural Language Processing and the 9th International Joint Conference on Natural Language Processing (EMNLP-IJCNLP)}.\hskip 1em plus 0.5em minus 0.4em\relax Hong Kong, China: Association for Computational Linguistics, Nov. 2019, pp. 3615--3620. [Online]. Available: \url{https://aclanthology.org/D19-1371}
\BIBentrySTDinterwordspacing

\bibitem{Liu_Zhou_Zhao_Wang_Ju_Deng_Wang_2020}
\BIBentryALTinterwordspacing
W.~Liu, P.~Zhou, Z.~Zhao, Z.~Wang, Q.~Ju, H.~Deng, and P.~Wang, ``K-bert: Enabling language representation with knowledge graph,'' \emph{Proceedings of the AAAI Conference on Artificial Intelligence}, vol.~34, no.~03, pp. 2901--2908, Apr. 2020. [Online]. Available: \url{https://ojs.aaai.org/index.php/AAAI/article/view/5681}
\BIBentrySTDinterwordspacing

\bibitem{vaswani2017attention}
A.~Vaswani, N.~Shazeer, N.~Parmar, J.~Uszkoreit, L.~Jones, A.~N. Gomez, {\L}.~Kaiser, and I.~Polosukhin, ``Attention is all you need,'' in \emph{Advances in neural information processing systems}, 2017, pp. 5998--6008.

\bibitem{liu2021pay}
H.~Liu, Z.~Dai, D.~So, and Q.~V. Le, ``Pay attention to mlps,'' \emph{Advances in neural information processing systems}, vol.~34, pp. 9204--9215, 2021.

\bibitem{sennrich2015neural}
R.~Sennrich, B.~Haddow, and A.~Birch, ``Neural machine translation of rare words with subword units,'' \emph{arXiv preprint arXiv:1508.07909}, 2015.

\bibitem{mikolov2013exploiting}
T.~Mikolov, Q.~V. Le, and I.~Sutskever, ``Exploiting similarities among languages for machine translation,'' \emph{arXiv preprint arXiv:1309.4168}, 2013.

\bibitem{moller2020covid}
T.~M{\"o}ller, A.~Reina, R.~Jayakumar, and M.~Pietsch, ``Covid-qa: A question answering dataset for covid-19,'' in \emph{Proceedings of the 1st Workshop on NLP for COVID-19 at ACL 2020}, 2020.

\bibitem{jin2019pubmedqa}
Q.~Jin, B.~Dhingra, Z.~Liu, W.~Cohen, and X.~Lu, ``Pubmedqa: A dataset for biomedical research question answering,'' in \emph{Proceedings of the 2019 Conference on Empirical Methods in Natural Language Processing and the 9th International Joint Conference on Natural Language Processing (EMNLP-IJCNLP)}, 2019, pp. 2567--2577.

\bibitem{yu2021dict}
W.~Yu, C.~Zhu, Y.~Fang, D.~Yu, S.~Wang, Y.~Xu, M.~Zeng, and M.~Jiang, ``Dict-bert: Enhancing language model pre-training with dictionary,'' \emph{arXiv preprint arXiv:2110.06490}, 2021.

\bibitem{bosselut2019comet}
A.~Bosselut, H.~Rashkin, M.~Sap, C.~Malaviya, A.~Celikyilmaz, and Y.~Choi, ``Comet: Commonsense transformers for automatic knowledge graph construction,'' in \emph{Proceedings of the 57th Annual Meeting of the Association for Computational Linguistics}, 2019, pp. 4762--4779.

\bibitem{poerner-etal-2020-e}
\BIBentryALTinterwordspacing
N.~Poerner, U.~Waltinger, and H.~Sch{\"u}tze, ``{E}-{BERT}: Efficient-yet-effective entity embeddings for {BERT},'' in \emph{Findings of the Association for Computational Linguistics: EMNLP 2020}.\hskip 1em plus 0.5em minus 0.4em\relax Online: Association for Computational Linguistics, Nov. 2020, pp. 803--818. [Online]. Available: \url{https://aclanthology.org/2020.findings-emnlp.71}
\BIBentrySTDinterwordspacing

\bibitem{sharma-etal-2019-incorporating}
\BIBentryALTinterwordspacing
S.~Sharma, B.~Santra, A.~Jana, S.~Tokala, N.~Ganguly, and P.~Goyal, ``Incorporating domain knowledge into medical {NLI} using knowledge graphs,'' in \emph{Proceedings of the 2019 Conference on Empirical Methods in Natural Language Processing and the 9th International Joint Conference on Natural Language Processing (EMNLP-IJCNLP)}.\hskip 1em plus 0.5em minus 0.4em\relax Hong Kong, China: Association for Computational Linguistics, Nov. 2019, pp. 6092--6097. [Online]. Available: \url{https://aclanthology.org/D19-1631}
\BIBentrySTDinterwordspacing

\bibitem{bodenreider2004unified}
O.~Bodenreider, ``The unified medical language system (umls): integrating biomedical terminology,'' \emph{Nucleic acids research}, vol.~32, no. suppl\_1, pp. D267--D270, 2004.

\bibitem{jin2019probing}
Q.~Jin, B.~Dhingra, W.~W. Cohen, and X.~Lu, ``Probing biomedical embeddings from language models,'' \emph{arXiv preprint arXiv:1904.02181}, 2019.

\bibitem{chen2017enhanced}
Q.~Chen, X.~Zhu, Z.-H. Ling, S.~Wei, H.~Jiang, and D.~Inkpen, ``Enhanced lstm for natural language inference,'' in \emph{Proceedings of the 55th Annual Meeting of the Association for Computational Linguistics (Volume 1: Long Papers)}, 2017, pp. 1657--1668.

\bibitem{xing2015normalized}
C.~Xing, D.~Wang, C.~Liu, and Y.~Lin, ``Normalized word embedding and orthogonal transform for bilingual word translation,'' in \emph{Proceedings of the 2015 Conference of the North American Chapter of the Association for Computational Linguistics: Human Language Technologies}, 2015, pp. 1006--1011.

\bibitem{zhang2019girls}
M.~Zhang, K.~Xu, K.-i. Kawarabayashi, S.~Jegelka, and J.~Boyd-Graber, ``Are girls neko or sh{\=o}jo? cross-lingual alignment of non-isomorphic embeddings with iterative normalization,'' in \emph{Proceedings of the 57th Annual Meeting of the Association for Computational Linguistics}, 2019, pp. 3180--3189.

\bibitem{rajpurkar-etal-2018-know}
\BIBentryALTinterwordspacing
P.~Rajpurkar, R.~Jia, and P.~Liang, ``Know what you don{'}t know: Unanswerable questions for {SQ}u{AD},'' in \emph{Proceedings of the 56th Annual Meeting of the Association for Computational Linguistics (Volume 2: Short Papers)}.\hskip 1em plus 0.5em minus 0.4em\relax Melbourne, Australia: Association for Computational Linguistics, Jul. 2018, pp. 784--789. [Online]. Available: \url{https://aclanthology.org/P18-2124}
\BIBentrySTDinterwordspacing

\bibitem{maldonado2019adversarial}
R.~Maldonado, M.~Yetisgen, and S.~M. Harabagiu, ``Adversarial learning of knowledge embeddings for the unified medical language system,'' \emph{AMIA Summits on Translational Science Proceedings}, vol. 2019, p. 543, 2019.

\bibitem{aronson2010overview}
A.~R. Aronson and F.-M. Lang, ``An overview of metamap: historical perspective and recent advances,'' \emph{Journal of the American Medical Informatics Association}, vol.~17, no.~3, pp. 229--236, 2010.

\bibitem{wallace2019nlp}
E.~Wallace, Y.~Wang, S.~Li, S.~Singh, and M.~Gardner, ``Do nlp models know numbers? probing numeracy in embeddings,'' in \emph{Proceedings of the 2019 Conference on Empirical Methods in Natural Language Processing and the 9th International Joint Conference on Natural Language Processing (EMNLP-IJCNLP)}, 2019, pp. 5307--5315.

\bibitem{hornik1989multilayer}
K.~Hornik, M.~Stinchcombe, and H.~White, ``Multilayer feedforward networks are universal approximators,'' \emph{Neural networks}, vol.~2, no.~5, pp. 359--366, 1989.

\bibitem{schick2020bertram}
T.~Schick and H.~Sch{\"u}tze, ``Bertram: Improved word embeddings have big impact on contextualized model performance,'' in \emph{Proceedings of the 58th Annual Meeting of the Association for Computational Linguistics}, 2020, pp. 3996--4007.

\bibitem{xu2020fusing}
Y.~Xu, C.~Zhu, R.~Xu, Y.~Liu, M.~Zeng, and X.~Huang, ``Fusing context into knowledge graph for commonsense reasoning,'' \emph{arXiv preprint arXiv:2012.04808}, 2020.

\bibitem{castelli-etal-2020-techqa}
V.~Castelli, R.~Chakravarti, S.~Dana, A.~Ferritto, R.~Florian, M.~Franz, D.~Garg, D.~Khandelwal, J.~S. McCarley, M.~McCawley \emph{et~al.}, ``The techqa dataset,'' in \emph{Proceedings of the 58th Annual Meeting of the Association for Computational Linguistics}, 2020, pp. 1269--1278.

\bibitem{soni2022radqa}
S.~Soni, M.~Gudala, A.~Pajouhi, and K.~Roberts, ``Radqa: A question answering dataset to improve comprehension of radiology reports,'' in \emph{Proceedings of the thirteenth language resources and evaluation conference}, 2022, pp. 6250--6259.

\end{thebibliography}

\end{document}